\documentclass{article} 
\usepackage{iclr2027_conference,times}

\usepackage{amsmath,amsfonts,bm}

\def\eqref#1{equation~\ref{#1}}

\def\1{\bm{1}}

\DeclareMathAlphabet{\mathsfit}{\encodingdefault}{\sfdefault}{m}{sl}
\SetMathAlphabet{\mathsfit}{bold}{\encodingdefault}{\sfdefault}{bx}{n}

\usepackage{amssymb}
\usepackage{hyperref}
\usepackage{url}
\usepackage{graphicx}
\usepackage{pgfplots}
\usepackage{tikz}
\usepackage{enumitem}
\usepackage{wrapfig}
\usetikzlibrary{patterns}
\usepgfplotslibrary{groupplots}
\usepackage{wrapfig}
\usepackage{booktabs}
\usepackage{listings}
\pgfplotsset{compat=1.18}

\definecolor{plaincolor}{RGB}{221,218,245}
\definecolor{numbercolor}{RGB}{221,236,213}
\definecolor{patterncolor}{RGB}{140,140,140}

\title{Diagnosing the Reference Gap: What Visual Scaffolds Can and Cannot Fix in Multimodal Reasoning}

\author{
  Guizhen Chen$^{1,2,}$\thanks{Guizhen is under the Joint PhD Program between Alibaba and NTU. Work done while Weiwen was at Alibaba.}
  \quad Weiwen Xu$^{2,}$\thanks{Corresponding authors.}
  \quad Wotao Yin$^{2}$
  \quad \textbf{Anh Tuan Luu}$^{1}$
  \quad Hou Pong Chan$^{3,4,}$\footnotemark[2]\\
  $^1$ Nanyang Technological University
  \quad $^2$ DAMO Academy, Alibaba Group US \\ 
  $^3$ Department of Artificial Intelligence, University of Macau \\
  $^4$ Institute of Artificial Intelligence and Brain Sciences, University of Macau\\
}

\iclrfinalcopy 
\begin{document}

\maketitle

\begin{abstract}
Multimodal large language models can recognize visual content yet struggle to reason reliably over it across multiple steps. In counting, a model may see all objects but lose track of which ones it has counted. We argue that such failures, commonly attributed to perception, reveal a reference gap: the failure to bind language to the relevant visual entities, regions, or locations. 
Motivated by the theory of visual routines, we test whether external references such as numbering can narrow this gap. We introduce ReGap-Bench, a paired benchmark of five visual reasoning tasks in which the scene, question, and answer are fixed and only the reference varies. Results show that task-matched references improve accuracy by up to 25\%, while mismatched references provide little benefit or even hurt performance. 
These findings identify the reference gap as a bottleneck distinct from visual perception and show that constructing task-appropriate visual references can serve as a training-free visual skill for improving multimodal reasoning.
Our data and code are available at \url{https://regap-bench.github.io}.
\end{abstract}

\section{Introduction}

\begin{figure}[t]
    \centering
    \includegraphics[width=0.95\linewidth]{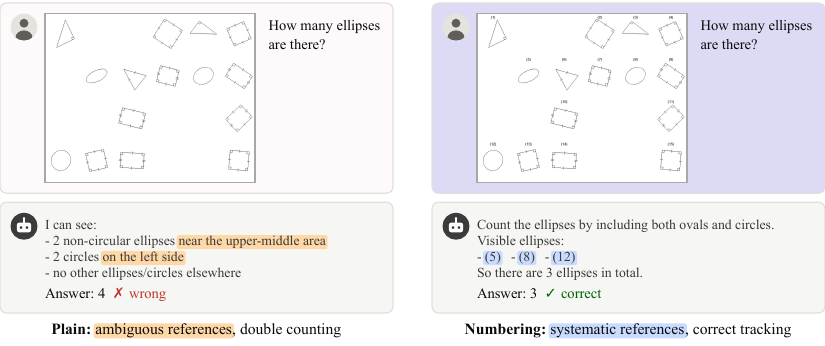}
    \caption{The same question to GPT-5.4 mini about the same scene, without and with numbering reference. In the plain setting (left), the model refers to shapes by vague locations and double counts. With numbering (right), it lists the ellipses by label and answers correctly.}
    \label{fig:teaser}
    \vspace{-0.15in}
\end{figure}

Multimodal large language models (MLLMs) have become increasingly capable at recognizing objects, shapes, and spatial relations in single-step perceptual tasks \citep{dong2026refadv,kimiteam2026kimik3openfrontier}, yet they often struggle to maintain and use that information across multiple reasoning steps. For example, a model may detect every object in a scene but still miscount them. Much of the field's response to such failures has focused on perception, pursuing higher-resolution encoders and finer-grained visual features~\citep{tong2024eyes}. However, the missing capability may instead lie between perception and reasoning: models lack a stable mechanism for referring to objects, regions, or locations while reasoning. Figure~\ref{fig:teaser} shows an example of a counting problem. The model refers to shapes by vague locations such as ``on the left side'' and counts one ellipse twice; when every shape carries a number, it lists the ellipses by label and answers correctly.

Following \citet{deepseek2026thinking}, we call this the \textit{reference gap}: the failure to maintain stable bindings between language-based reasoning and the relevant visual entities, regions, or locations. Recent work addresses it either by training a model to produce coordinates within its reasoning \citep{deepseek2026thinking}, or by drawing lines on the image and instructing the model to scan along them \citep{izadi2025visual}. Two questions remain open. First, neither isolates the effect of the reference itself. \citet{deepseek2026thinking} compare a newly trained model with other models, so part of the gain may come from training data or architecture. \citet{izadi2025visual} varies both the image and the prompt together, so it is unclear if MLLMs can perform systematic scanning of the image without additional instructions from the prompt. Second, each applies the same reference to every task, although tasks might need different visual operations. It thus remains unclear which reference a task needs, and whether any added structure would help equally.

To design such a diagnostic framework, we identify visual routines used for complex visual reasoning, such as marking visited items, activating bounded regions, and tracing boundaries \citep{ullman1984visual}. We hypothesize that MLLMs lack internal reference frames for reliable execution of these visual routines, and that external references such as numbering, shading, or grid lines can support them, with each reference helping most on tasks that demand its visual routines.
To test this hypothesis, we introduce \textbf{ReGap-Bench}, a controlled benchmark of paired reference interventions across five tasks: shape counting, angle ranking, size ranking, clock ranking, and transformation prediction. Each underlying scene is rendered with and without explicit visual references, while its task-relevant objects, geometry, question, and gold answer remain fixed. 

Results show that task-matched references improve accuracy on all five tasks. Numbering yields gains of up to 17.5\% on shape counting and 25\% clock ranking; shading improves angle and size ranking; and grids improve size ranking and transformation prediction. Next, we show that the gains depend on matching the reference to the task, not on adding marks. Additionally, model scale across six Qwen3-VL models from 2B to 235B does not remove the reference gap. 
Building on these findings, 
we introduce a \textbf{training-free visual skill} that enables an agent to dynamically identify the required visual routine and select the corresponding reference scaffold. Evaluated on CartoMapQA \citep{CartoMapQA2025Ung}, this skill enables autonomous reference selection and improves complex map understanding without finetuning.
We summarize our contributions as follows:
\begin{itemize}[leftmargin=*,topsep=2pt]
\setlength{\itemsep}{2pt}
\setlength{\parskip}{0pt}
\setlength{\parsep}{0pt}
    \item We introduce ReGap-Bench, a paired evaluation framework that measures the reference gap. It spans five visual reasoning tasks and three reference systems, with scene semantics, questions, and answers held fixed across reference conditions.
    \item We show that task-matched references improve all evaluated models on all five tasks. Mismatched, incomplete, and model-scale analyses provide evidence that the gains come from supporting the visual routine a task requires, and that they persist from 2B to 235B parameters.
    \item We introduce a training-free visual skill that enables an agent to identify the visual routine required by a task and dynamically construct a corresponding visual reference at inference time. 
\end{itemize}

\section{Diagnosing the Reference Gap with Visual Interventions}
\subsection{The reference gap}
\label{sec:reference-gap}

We use the reference gap to describe the failure to maintain a stable link between language and the relevant visual entity, region, or location that the language corresponds to. Models might perceive the relevant visual information, yet struggle to identify, revisit, or distinguish from other visual elements as reasoning proceeds. Consider shape counting. A model may correctly recognize individual triangles when asked about them separately yet still miscount them (see preliminary analysis in Appendix \ref{app:perception}). The model needs to repeatedly select an object, determine whether it belongs to the target class, keep track of counted objects, and then move to the next. Without a stable internal reference, it can lose track of which objects have been counted, resulting in omissions or duplicates. The reference gap is therefore distinct from a perception failure, and it should matter most in visual reasoning that spans multiple steps.

Most benchmarks do not reveal whether the model failed to perceive the relevant content or failed to maintain reference to it during reasoning. To address this, we diagnose the reference gap by testing whether performance changes when the underlying visual problem is held fixed but the model is given explicitly external reference to refer to its visual elements.

\subsection{External references for visual routines}

To choose useful visual references, we draw on Ullman's theory of visual routines ~\citep{ullman1984visual}, which describes complex visual reasoning as a sequence of elementary operations such as shifting the processing focus, bounded activation, boundary tracing, indexing, and marking. We propose three types of visual references that externalize intermediate visual states to support these routines.

\textbf{Numbering.} Numbering supports marking and indexing. It turns description-dependent visual elements into addressable entities. For instance, instead of using ambiguous phrases like ``the upper-left triangle", the model can refer to ``shape 7''. This makes it easier to find an item again and to track which items have already been processed.

\textbf{Shading.} Shading supports bounded activation by making a region explicit. For example, shading the interior of an angle makes the angular region itself visible. Filling a shape similarly emphasizes its bounded area rather than only its outline. 

\textbf{Grid lines.} Grids provide spatial marks that can support systematic focus shifts and boundary tracing. They divide the image into regular grid intervals and can provide a reference for position, width, height, displacement, and alignment.

These scaffolds provide different kinds of reference information. Therefore, we do not expect every scaffold to help every task. Our main hypothesis is that: a reference scaffold helps most when the type of reference it provides matches the visual routines that the reasoning process needs.

\subsection{Paired reference-intervention design}
\label{sec:intervention}

To test whether a reference gap exists, we design a paired reference intervention in which we hold the problem fixed and intervene only on the external reference available to the model. For each problem, we create a plain image and one or more versions with added references, all generated from the same underlying scene.

Formally, a problem is a triplet $(s, q, y)$, where $s$ is the scene specifying the objects, geometry, and relations; $q$ is the question; and $y = f(s, q)$ is the ground truth determined by the scene and question. 

Let $\mathcal{C}$ be a set of reference conditions including the plain condition $c_0$, and let $x_c = R(s, c)$ denote the image obtained by rendering scene $s$ under condition $c \in \mathcal{C}$. Every reference condition satisfies three properties:
(i) \emph{content preservation}: every element of $s$ appears in $x_c$ with the same shape, size, and position as in $x_{c_0}$, and the reference only adds numbers, changes fill color, or adds background structure;
(ii) \emph{answer invariance}: the target $y = f(s, q)$ is identical for all $c \in \mathcal{C}$ as it only depends on $s$ and $q$; and
(iii) \emph{no answer leakage}: the reference alone does not determine $y$, since numbers and colors are assigned to all elements independently of their class.
Each problem therefore yields a set of images $\{x_c\}_{c \in \mathcal{C}}$ that differ only in the reference available to the model.

Given a model $M$, let $\hat{y}_c = M(x_c, q)$ denote its prediction under condition $c$, and let $a_c = \mathbf{1}[\hat{y}_c = y]$ denote its correctness. For a reference condition $c \neq c_0$, the effect of adding the reference is $\delta(c) = a_c - a_{c_0} \in \{-1, 0, 1\}$. For a dataset $\mathcal{D} = \{(s_i, q_i, y_i)\}_{i=1}^{N}$ rendered under condition $c$, the average reference effect is:
\begin{equation}
  \Delta(c) = \frac{1}{N} \sum_{i=1}^{N} \delta_i(c) = \mathrm{Acc}(c) - \mathrm{Acc}(c_0),
  \qquad
  \mathrm{Acc}(c) = \frac{1}{N} \sum_{i=1}^{N} a_{i,c}.
\end{equation}

A positive $\Delta(c)$ suggests a reference gap: the model is able to solve more of the same problems once the reference is made explicit. Since adding marks also introduces additional visual information, we further test in Section \ref{sec:mismatch} whether the gains come from the reference itself. A close-to-zero $\Delta(c)$ means the added reference brings no benefit, either because the model already maintains reference on its own or because its errors lie somewhere else, such as in perception. A negative $\Delta(c)$ means the reference interferes with the task, for example by adding clutter to the scene.

\section{\textbf{ReGap-Bench}: A Reference-Intervention Benchmark}
\label{sec:bench}

\begin{figure}[t]
    \centering
    \includegraphics[width=0.95\linewidth]{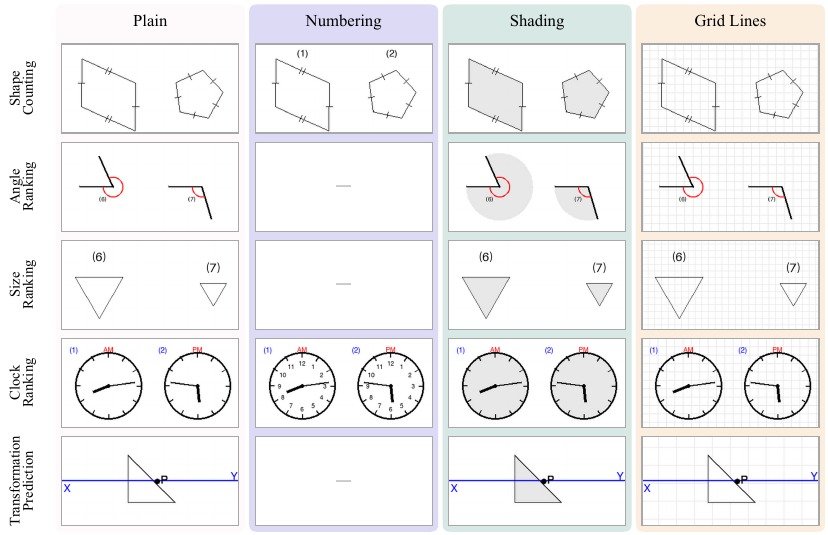}
    \caption{Example ReGap-Bench scenes. Each row shows a crop of one scene under the plain and each reference condition. The scene content, question, and answer are identical across a row. Numbering is not applied to angle ranking, size ranking, or transformation prediction.} 
    \label{fig:bench}
    \vspace{-0.1in}
\end{figure}

We introduce \textbf{ReGap-Bench}, applying the paired design of Section~\ref{sec:intervention} to five tasks: shape counting, angle ranking, size ranking, clock ranking, and transformation prediction. Each task has 200 scenes, each rendered to different reference conditions, as shown in Figure~\ref{fig:bench}. A reference only adds numbers, fills interiors with light gray, or adds faint grid lines. It does not move, remove, or cover any original  (\emph{content preservation}), and it treats all elements alike, so it does not reveal the answer (\emph{no answer leakage}). Appendix~\ref{app:bench} shows generation details and dataset statistics.

\subsection{Tasks}
We design five tasks that share a common requirement: the model must shift its focus across different parts of an image and repeatedly bind reasoning steps to particular visual elements. For each task, we identify the visual routine it relies on most and the reference scaffolds we expect to support that routine, which we call \emph{matched}.

\textbf{Shape counting.} The scene contains outlined shapes such as triangles, quadrilaterals, and pentagons, arranged in rows and columns at random orientations. We annotate the shapes with standard geometry marks, such as tick marks on equal sides and small squares at right angles, to make each shape's class unambiguous. The question asks for the count of a specific shape. The model must select each shape, classify it, and record it as counted without skipping or repeating any. The task therefore relies on indexing and marking. Numbering could be a matched reference because it assigns each object a stable address that can be used throughout the counting process.

\textbf{Angle ranking.}
The scene contains several angles, each formed by two rays from a vertex, and marked by a red arc. The model is tasked to rank the angles by size, from smallest to largest, or the reverse. The main difficulty is keeping the correct angular region active: looking at the wrong side of the rays turns an angle into its conjugate angle. Shading the angular wedge could be helpful in making the region being compared explicit and thus supports bounded activation.

\textbf{Size ranking.}
The scene contains several shapes of one class. The model is asked to rank the shapes according to their visual size. This requires estimating the size of each shape and maintain these estimates while shifting the focus to other shapes. We expect two references to help: shading makes each shape's area explicit as a bounded region, and grid lines provide a common scale for comparing width and height across locations.

\textbf{Clock ranking.}
The scene contains several analog clocks, each with twelve tick marks, an hour hand, a minute hand, and an ``AM'' or ``PM'' label. The task is to rank the clocks from earliest to latest, or the reverse. The main reference challenge is mapping hand positions onto a stable circular reference frame. We expect dial numbering to help, because it marks positions on that frame directly.

\textbf{Transformation prediction.} Given a source figure and a chain of transformation steps, the model needs to determine which of the four given options is the final figure. Transformation operations include translation, reflection, rotation, and dilation. To solve the task, the model must trace the shape through each step and shift focus across panels to compare final positions. 
We expect grid lines to help, because they provide a common frame for position and scale.

\subsection{Evaluation Setup}

We evaluate five MLLMs from both proprietary and open-source families: Claude-Opus-4.7~\citep{anthropic2026claudeopus}, Gemini-3.1-Flash-Lite~\citep{google2026gemini31flashlite}, GPT-5.4-mini~\citep{openai2026gpt54mini}, Qwen3-VL-30B-A3B-Thinking~\citep{DBLP:journals/corr/qwen3vl}, and InternVL3.5-38B~\citep{DBLP:journals/corr/intervl35}. In Section \ref{sec:exp}, we refer to models by their family names (e.g., Gemini). To study the effect of model scale, we additionally evaluate six Qwen3-VL Thinking models, from 2B to 235B parameters\footnote{https://huggingface.co/collections/Qwen/qwen3-vl}. Prompt template is shown in Appendix~\ref{app:prompts}. Our main metric is exact match. For ranking tasks, we additionally report pairwise accuracy, the share of pairs that the predicted ranking puts in the correct order, which gives partial credit. We compute it over valid rankings.

\section{Experiment}\label{sec:exp}
\subsection{Task-matched references improve visual reasoning}
We begin with the operation-matched setting: for each reference, we test two tasks whose required visual routine could be supported by that scaffold. 

\begin{figure}[t]
    \centering
    \includegraphics[width=\linewidth]{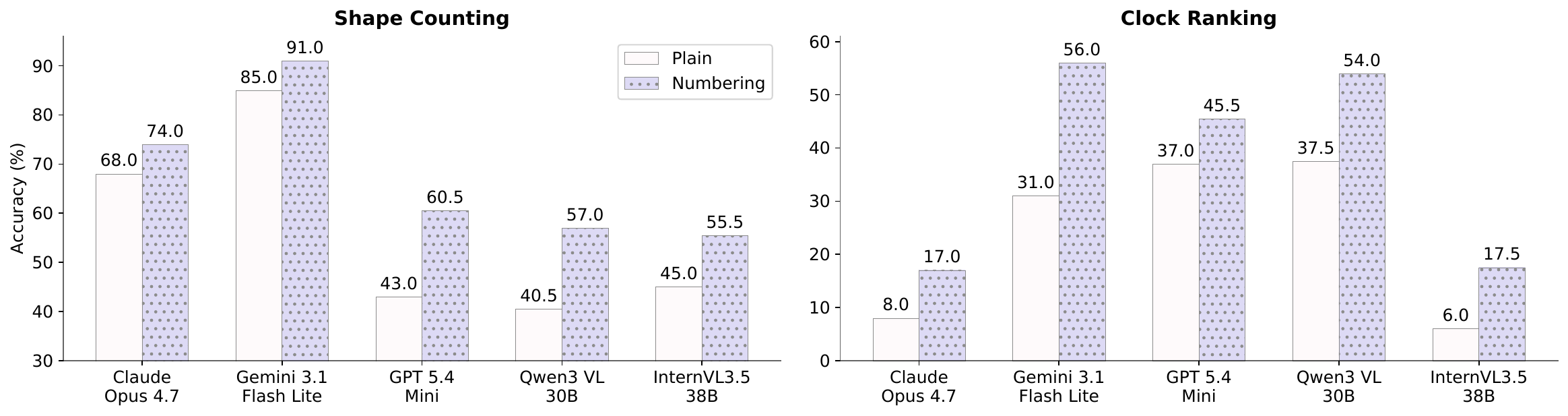}
    \caption{Numbering improves performance across all models on shape counting and clock ranking.}
    \label{fig:numbering}
    \vspace{-0.1in}
\end{figure}

\begin{figure}[t]
    \centering
    \includegraphics[width=\linewidth]{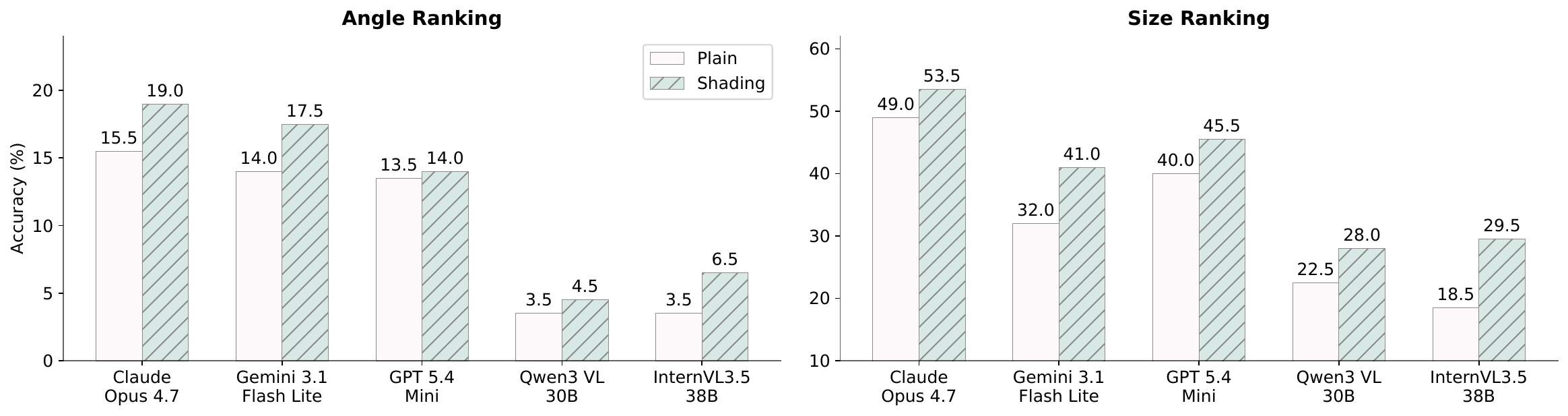}
    \caption{Shading improves across all evaluated models on both angle ranking and size ranking.}
    \label{fig:shading}
    \vspace{-0.1in}
\end{figure}
\paragraph{Numbering supports indexing and marking.}
Figure \ref{fig:numbering} shows that numbering consistently improves performance, with a performance gain of 6.0\% to 17.5\% on shape counting, and 8.5\% to 25.0\% on clock ranking. 
On shape counting, models with low accuracy in the plain condition gain the most: GPT and Qwen gain about 17\%, while Gemini and Claude gain 6\%. To examine how numbering changes the counting process, we manually label each model response by the reference strategy it uses: (i) \emph{systematic references}, which use numbered identifiers (e.g., ``shape (7)'') and a explicit row- or column-based scheme (e.g., ``Top row''); and (ii) \emph{ambiguous references}, which use vague spatial descriptions (e.g., ``the upper-left one'') or provide no clear reference scheme. 
On plain images, Gemini frequently constructs its own reference system: 52.0\% of its responses assign numbers to the shapes, and 31.5\% explicitly scan them by rows or columns. 
In contrast, GPT does so much less often: only 15.5\% of its responses use either \emph{systematic references} strategy, and the other 84.5\% use \emph{ambiguous references}. With numbering, 100\% of Gemini's and 86.0\% of GPT's responses refer to the printed numbers. This suggests that \textbf{numbering supplies stable symbolic addresses that are particularly useful for models that lack their own reference system}.

For clock ranking, we extract the time each model stated for each clock from its response and count a clock as correctly-read if it is within 5 minutes of the ground truth. Among clocks with a stated time, numbering raises this rate from 10.6\% to 28.6\% for Claude, 58.2\% to 82.5\% for Gemini, 67.5\% to 75.6\% for Qwen3-VL, and 2.9\% to 17.4\% for InternVL. 
Thus, \textbf{dial numbers improve these models' ability to read individual clocks, an intermediate step required for correct ranking}.
GPT behaves slightly differently: when it states a time, it is almost accurate (90.3\% plain vs.\ 90.8\% with numbering), but on plain images it gives no time for 34\% of the clocks, and this propotion is reduced to 12\% with numbering. For GPT, therefore, \textbf{numbering primarily improves the completeness of the intermediate clock-reading process rather than the accuracy of readings}.

\begin{wraptable}{r}{0.5\textwidth}
\vspace{-0.2in}
\caption{Size-ranking pairwise accuracy (\%, plain $\rightarrow$ shading) by scale-factor difference.}
\label{tab:app-sizegap}
\centering
\small
\setlength{\tabcolsep}{3pt}
\resizebox{\linewidth}{!}{%
\begin{tabular}{lccc}
\toprule
& \multicolumn{3}{c}{\textbf{Scale-factor difference}} \\
\cmidrule(lr){2-4}
\textbf{Model} & \textbf{0.05--0.10} & \textbf{0.10--0.20} & \textbf{$\geq$ 0.20} \\
\midrule
Claude   & 82.2 $\rightarrow$ 84.3 & 95.6 $\rightarrow$ 96.6 & 99.8 $\rightarrow$ 99.9 \\
Gemini   & 79.3 $\rightarrow$ 82.4 & 91.8 $\rightarrow$ 93.1 & 99.3 $\rightarrow$ 99.5 \\
GPT      & 80.0 $\rightarrow$ 80.8 & 93.4 $\rightarrow$ 94.4 & 99.5 $\rightarrow$ 99.6 \\
Qwen3-VL & 71.6 $\rightarrow$ 74.0 & 84.3 $\rightarrow$ 85.7 & 97.6 $\rightarrow$ 98.2 \\
InternVL & 67.5 $\rightarrow$ 73.6 & 84.2 $\rightarrow$ 86.8 & 97.5 $\rightarrow$ 98.7 \\
\bottomrule
\end{tabular}%
}
\vspace{-0.1in}
\end{wraptable}
\paragraph{Shading supports bounded activation.}
Figure \ref{fig:shading} shows the effects of shading on angle ranking and size ranking. Shading improves size ranking from 4.0\% to 11.0\%, while the gains on angle ranking are relatively small. As exact-match is a very strict metric for this task, and plain accuracy is all below 16\%, we additionally study the pairwise accuracy (Appendix \ref{app:results}).  Pairwise accuracy improves with shading for four of the five models, including from 76.4\% to 85.2\% for Gemini and from 59.7\% to 68.0\% for Qwen3-VL. Thus, \textbf{shading improves the overall ordering of the angles significantly}. 

On size ranking, pairwise accuracy is already above 91\% for every model on plain images, and the remaining errors lie almost entirely between shapes of similar size: pairs whose scales differ by less than 0.10 are ordered correctly only 67.5\% to 82.2\% of the time, compared with at least 97.5\% for pairs that differ by 0.20 or more (Table~\ref{tab:app-sizegap}). Shading helps most on those close pairs, suggesting that \textit{the exact-matching gains primarily come from resolving swaps between similarly sized shapes}.

To study how shading helps angle ranking, we measure one key error that it likely prevents: reading the wrong side of rays, which turns an angle $\theta$, into its conjugate angle, $360^\circ - \theta$. For instance, Claude reads one angle as about $10^\circ$ on the plain image and as about $350^\circ$ with shading, where the true value is $351^\circ$. As models do not always state a degree value for each angle, we measure this error from the final rankings. We take each angle below $90^\circ$ or above $270^\circ$, for which the two readings are far apart, and check whether the model placed it closer to where $\theta$ belongs or to where $360^\circ - \theta$ would belong. Detailed implementation can be found in Appendix~\ref{app:wrong-side}.
With shading, it almost halves this rate for Claude (3.3\% to 1.7\%), Gemini (18.4\% to 8.5\%), and InternVL (23.9\% to 13.1\%). This shows that \textbf{shading provides a bounded visual region that helps models maintain the intended side of the rays during measurement}. GPT-5.4-mini shows no such reduction (13.3\% vs. 13.8\%), consistent with its nearly unchanged pairwise accuracy under shading.

\begin{figure}[t]
    \centering
    \includegraphics[width=\linewidth]{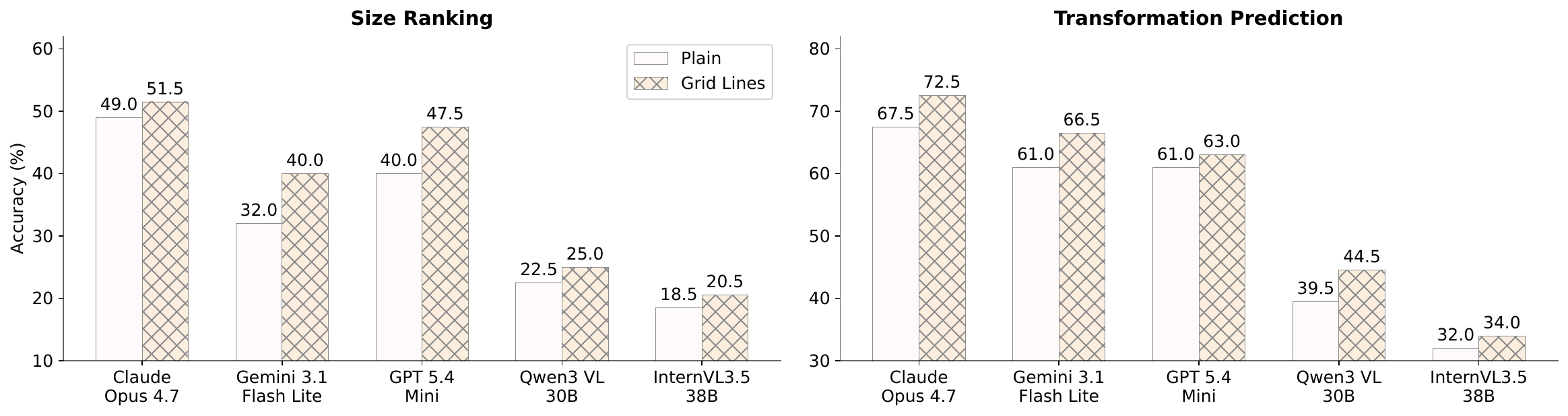}
    \caption{Grid lines improve all evaluated models on size ranking and transformation prediction.}
    \label{fig:grid}
    \vspace{-0.1in}
\end{figure}

\paragraph{Grids support boundary tracing and shifting focus.}
Figure \ref{fig:grid} shows that grid lines consistently improve performance, with a performance gain of 2.0\% to 8.0\% on size ranking, and 2.0\% to 5.5\% on transformation prediction. Notably, in size ranking, grid lines yields smaller gains than shading in four models. One possible explanation is that using grids requires several additional operations: the model must measure each shape in grid cells, keep each reading, and then compare the readings. These operations themselves depend on maintaining stable references. On the other hand, shading makes the region of each shape directly visible.

To examine whether models indeed use grids for measurement or position tracking, we use an LLM judge to classify the model responses by the reference strategy :
(i) \emph{systematic references} that give at least one position, distance, or size in grid cells or as grid coordinates; (ii) \emph{ambiguous references} that describe position and size only in general terms, which is detailed in Appendix~\ref{app:grid-judge}.
Gemini uses \emph{systematic references} in 51\% of its responses in transformation prediction, usually by writing the vertices as grid coordinates and following them through each step (e.g., ``The vertices are roughly at $(-1, 0)$, $(-1, 2)$, and $(0, 0)$ relative to P''). In contrast, GPT never adopts this strategy: it relies on qualitative descriptions such as ``left of P'', ``half the size'', or ``closer to P'', regardless of plain or grid condition. This difference is consistent with the larger improvement from grids for Gemini than for GPT (5.5\% vs. 2.0\%), suggesting that \textbf{the benefit of a reference scaffold depends not only on its presence, but also on whether the model incorporates it into its reasoning process}.


\begin{figure}[t]
    \centering
    \includegraphics[width=\linewidth]{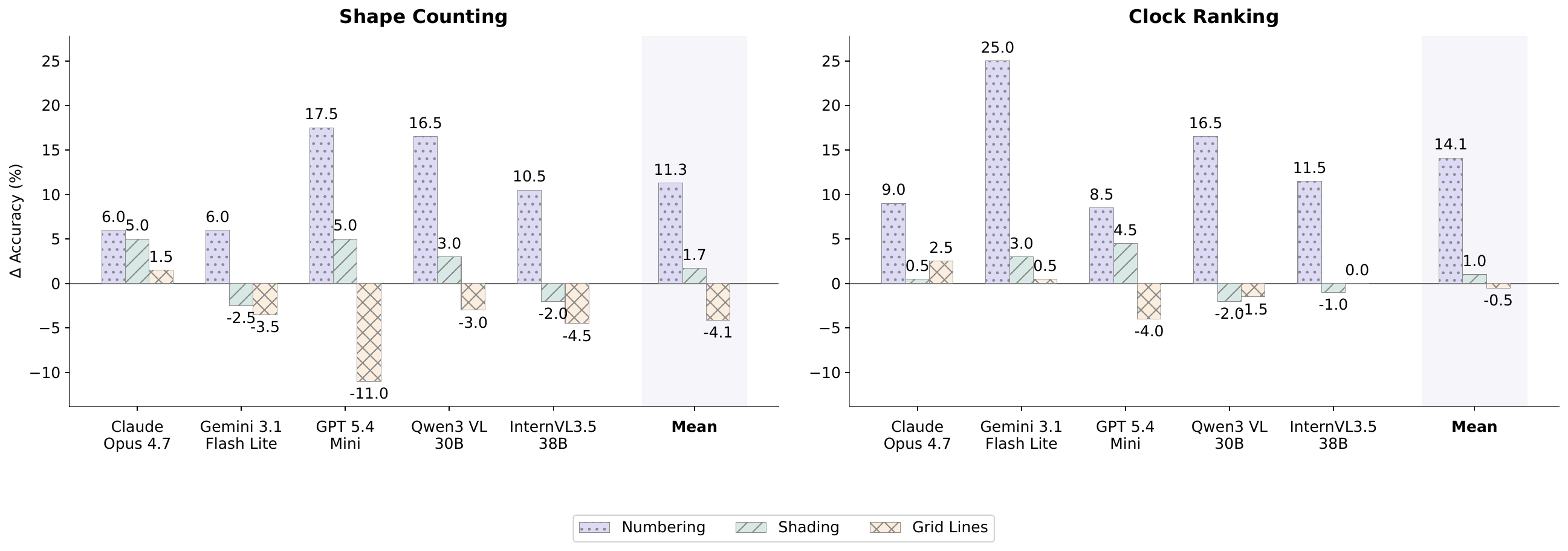}
    \caption{Operation-matched numbering produces consistently larger gains than shading or grid lines. Shading provides only small or inconsistent gains, while grid lines are often neutral or detrimental, suggesting that visual references help when they support the routines required by the task.}
    \label{fig:mismatch}
    \vspace{-0.1in}
\end{figure}

\subsection{Analysis on Matched Versus Mismatched References}
\label{sec:mismatch}

The positive scaffold results establish that each reference method can help in matched settings. However, it is unclear whether these gains simply come from adding more visual marks. We test this by applying numbering, shading, and grid lines to the same shape counting and clock ranking scenes. If improvements were caused merely by additional visual information, then numbering, shading, and grid lines should help similarly. If the reference gap exists, then the operation-matched scaffold should produce the largest gain. Figure \ref{fig:mismatch} supports the second explanation. In shape counting, numbering improves accuracy by 11.3\% on average, compared to only 1.7\% for shading, while grid lines reduce accuracy by 4.1\%. The same separation appears in clock ranking.
These results show that added visual structure is not always beneficial: a scaffold can be neutral or harmful when it does not provide the kind of reference the task needs. \textbf{The effectiveness of a scaffold therefore depends on its alignment with the visual routine needed for solving the task.}

\subsection{Analysis on Incomplete References}

\begin{wrapfigure}{r}{0.52\textwidth}
    \vspace{-0.15in}
    \centering
    \includegraphics[width=\linewidth]{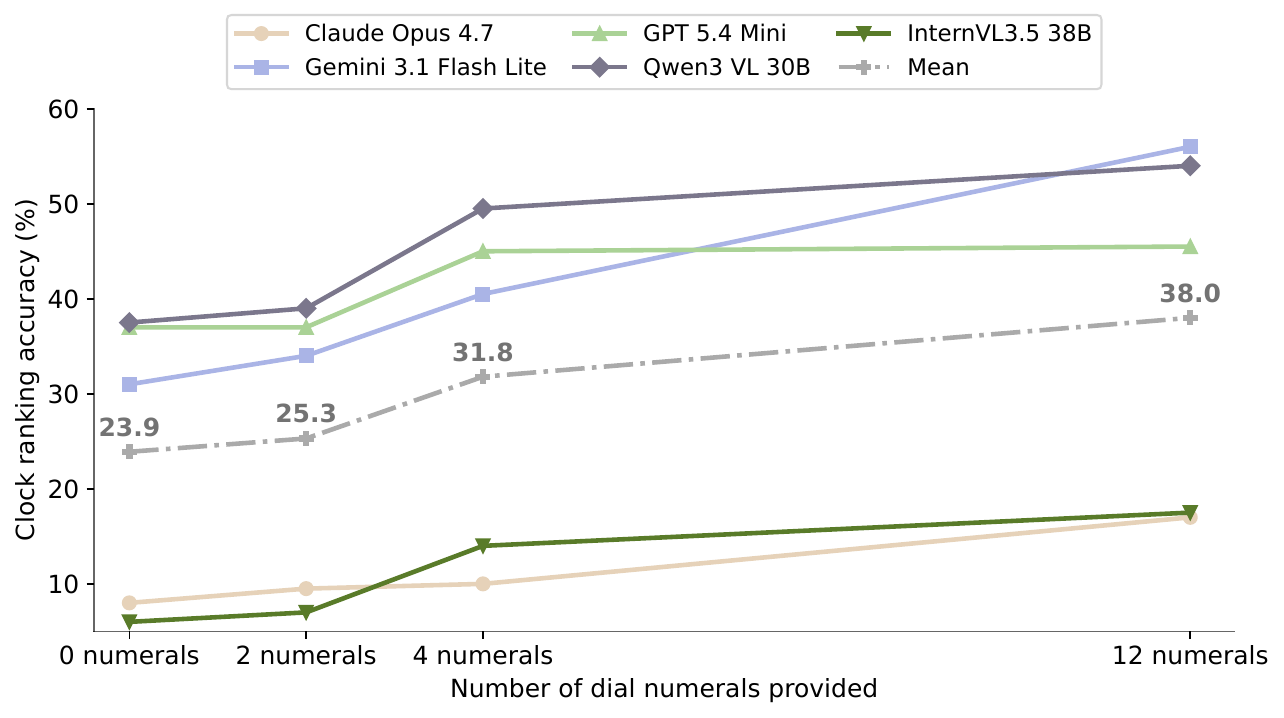}
    \caption{Performance improves as dial numbering becomes more complete.}
    \label{fig:incomplete}
\end{wrapfigure}

Next, we test whether references help because they provide complete reference structure, or whether partial reference is sufficient. We test this on clock ranking by varying how many dial numerals are shown while keeping the clocks themselves fixed. The four conditions contain zero, two, four, or all twelve numerals. If numbering helps because it provides stable addresses, then a more complete reference system should be more useful than a sparse one.  
Figure~\ref{fig:incomplete} confirms this, mean accuracy rises from 23.9\% with no numerals to 25.3\% with two, 31.8\% with four, and 38.0\% with all twelve. The trend is non-decreasing for every evaluated model. Sparse numbering therefore provides only a small benefit, whereas a complete reference system produces a substantially larger gain. This result is consistent with a reference-based mechanism: a scaffold is most useful when it supplies stable addresses that the model must repeatedly identify during reasoning. \textbf{Partial references can still help, but they leave a larger portion of the task dependent on the model's own spatial descriptions or implicit object bindings. }

\subsection{Analysis of Model Scale}
\begin{wrapfigure}{r}{0.52\textwidth}
    \vspace{-0.15in}
    \centering
    \includegraphics[width=\linewidth]{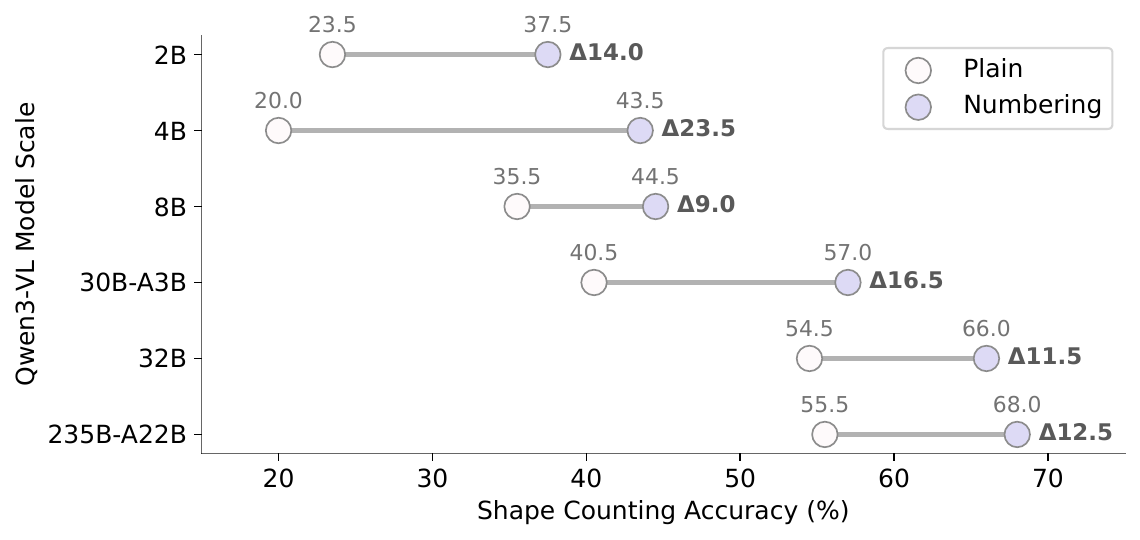}
    \caption{Scaling model size help both plain and numbering, but does not eliminate the reference gap.}
    \label{fig:scaling}
    \vspace{-5pt}
\end{wrapfigure}

The final question is whether the reference gap is primarily a limitation of small models. We evaluate six Qwen3-VL Thinking models ranging from 2B to 235B parameters on shape counting. Figure~\ref{fig:scaling} shows a clear overall increase in accuracy with model scale. However, the advantage of numbering remains at every scale: the gain ranges from 9.0\% to 23.5\% and does not decrease monotonically as models become larger. Thus, \textbf{a larger model size improves the task performance but does not eliminate the benefit of an explicit reference system}. 

\section{Task-matched references as training-free visual skills}

In Section \ref{sec:exp}, we show that models that lack an internal reference system can benefit from task-matched references \textit{without any training}. One implication is that agents can add these references on their own using a visual skill when a task requires such reference. Unlike most skills that are text-only, we package the three references as a visual skill that can refer to visual examples and overlay the reference onto the image similarly.

We test the skill on CartoMapQA~\citep{CartoMapQA2025Ung} that contains street-level OpenStreetMap renderings with icons for points of interest (POIs), road labels, markers, and a scale bar. We use stratified sampling to draw 100 tasks across the eight subtasks, and run the Codex GPT-5.5 agent on the same tasks with and without the skill, using the BenchFlow framework~\citep{benchflow2026}.

The skill contains three components: First, written rules specify which reference to use for each type of visual operation: numbering for counting, listing, or matching discrete map elements, as well as tracking junctions along a route; grid lines, calibrated using the map's scale bar, for measuring distances or comparing positions; shading for isolating bounded regions; and no overlay when the answer can be obtained from a single clearly labeled element. 
Second, a pair of before-and-after example images demonstrating how to draw each reference. Third, a script draws the chosen reference on a copy of the map and leaves the original unchanged.

\begin{wraptable}{r}{0.5\textwidth}
\vspace{-0.2in}
\caption{Visual skill performance (accuracy, \%) on CartoMapQA, broken down by overlay type.}
\label{tab:skills}
\centering
\small
\setlength{\tabcolsep}{4pt}
\begin{tabular}{lcccc}
\toprule
\textbf{Overlay} & \textbf{Tasks} & \textbf{No skill} & \textbf{With skill} & $\mathbf{\Delta}$ \\
\midrule
None       & 43 & 86.0 & 86.0 & 0.0 \\
Numbering  & 41 & 63.7 & 70.1 & +6.4 \\
Grid lines & 13 & 92.5 & 96.7 & +4.2 \\
Shading    &  3 & 33.3 & 66.7 & +33.3 \\
\bottomrule
\end{tabular}
\vspace{-0.1in}
\end{wraptable}
Table~\ref{tab:skills} shows that all references improve the mean score. The agent draws an overlay on 57 of the 100 tasks, and the selected references broadly align with the visual operations required by each subtask: numbering for counting, listing, and route navigation; grid lines for route-length tasks; shading for region isolation for listing tasks; and no overlay for symbol recognition and presence questions. These results suggest that \textbf{an agent can select and construct task-relevant references through a reusable skill without additional model training}.


\section{Related Work}
\textbf{Visual perception and reasoning in MLLMs.}
Benchmarks including BLINK \citep{fu2024blink}, MMVP \citep{tong2024eyes}, BlindTest \citep{rahmanzadehgervi2024blind}, CV-Bench \citep{tong2024cambrian}, and V*Bench \citep{wu2024vstar} study limitations in visual perception and reasoning. Related studies report difficulties in spatial reasoning \citep{wang2024spatialeval,yang2024thinkinginspace}, enumeration \citep{tamarapalli2025countqa,deng2026holocount}, and chart understanding and reasoning \citep{masry2025chartqapro}. However, they do not distinguish failures to encode visual elements from failures to maintain references during reasoning. Our paired design probes the effects of visual references by holding the underlying scene, question, and answer fixed while varying the reference system. 

\textbf{Grounding intermediate reasoning in visual references.}
\citet{deepseek2026thinking} identify the \emph{reference gap} as a failure to ground verbal reasoning in precise visual entities and locations. Such failures can propagate across reasoning steps in dense counting, spatial deduction, and path tracing. Their approach trains models to generate points and bounding boxes within their reasoning trajectories through large-scale primitive supervision, reinforcement learning, and distillation. BLINK highlights tasks that are difficult to mediate through natural language \citep{fu2024blink}. On VSI-Bench, \citet{yang2024thinkinginspace} find that chain-of-thought prompting, self-consistency, and tree-of-thoughts reduce Gemini-1.5 Pro~\cite{team2024gemini15pro}'s average performance in their evaluation, while explicitly generating cognitive maps improves its relative-distance accuracy. 
We examine whether references supplied in the input image can support these bindings in a fixed, off-the-shelf model, and under what conditions the model uses them.

\textbf{Visual prompting and grounded reasoning.}
Prior work supports visual reasoning through annotations, auxiliary visual states, and modular computation. Red-circle prompting directs CLIP towards selected regions \citep{shtedritski2023redcircle}, while Set-of-Mark supplies identifiers for segmented regions \citep{yang2023setofmark}. Visual Sketchpad generates auxiliary drawings \citep{hu2024visualsketchpad}, PIVOT refines action proposals through numbered annotations \citep{nasiriany2024pivot}, and Draw-and-Understand trains models to interpret user-drawn prompts \citep{lin2025drawandunderstand}. VisAidMath evaluates mathematical reasoning with textually described visual aids \citep{ma-etal-2026-visaidmath}. Other approaches compose vision modules through code \citep{suris2023vipergpt}, incorporate visual features into rationale generation \citep{zhang2023multimodalcot}, or represent intermediate reasoning states as images \citep{meng2024coi}. Our study systematically varying references rendered in the input image to test which reference systems support particular visual routines and whether their benefits depend on task alignment, with task-matched references implemented as reusable skills that require no model training.

\section{Conclusion}
We identify the \textit{reference gap} as a distinct bottleneck in multimodal reasoning: models may perceive the relevant visual content yet fail to maintain stable references to it during multi-step reasoning. Using ReGap-Bench, we show that task-matched visual references consistently improve performance, whereas mismatched references provide little benefit or can be harmful, and that this gap persists even at larger model scales. These results suggest that improving multimodal reasoning requires 
mechanisms that support the visual routines needed by the task. Finally, our training-free visual skill demonstrates that an agent can dynamically construct such references at inference time, offering a simple way to improve visual reasoning without additional model training.

\subsection*{AI use statement}
In this work, we used generative AI tools to refine hypotheses and support qualitative analysis. 
We have not used generative AI tools to generate synthetic data sets, clean or reformat dataset.
Lastly, helping develop theoretical models or conceptual frameworks, formulating mathematical claims, and proving mathematical claims are not applicable to this work.

Additionally, we used generative AI tools to edit software code, polish a research paper to improve readability, and identify relevant literature. We have reviewed all AI-assisted work. For example, we carefully reviewed the AI-refined paper content and evaluated AI-assisted implementation using manually written unit tests. We take responsibility for the final content of this work, including text, claims or artifacts produced with the aid of generative AI.

\section*{Acknowledgment}
This work was jointly supported by the Multi-year Research Grant from the University of Macau (Grant No. MYRG-SRG2026-00032-FIC) and DAMO Academy
through Alibaba-NTU Singapore Joint Research
Institute (JRI). We would also like to extend our gratitude to Interdisciplinary Graduate Programme and
College of Computing and Data Science of NTU for their support.

\bibliography{iclr2027_conference}
\bibliographystyle{iclr2027_conference}

\clearpage
\appendix
\section{Preliminary Check: Can Models Perceive the Shapes?}
\label{app:perception}

Section~\ref{sec:reference-gap} claims that a model can perceive the shapes in a scene yet still fail to count a target class. As a preliminary check of this claim, we run two simpler probes on the plain shape-counting scenes with Claude-Opus-4.7.

\paragraph{Finding the shapes.} We ask the model how many shapes the image contains, regardless of class. It answers correctly in 193 of the 200 scenes (96.5\%; Table~\ref{tab:app-total}). The errors grow with the number of shapes.

\begin{table}[h]
\caption{Counting all shapes, regardless of class (Claude-Opus-4.7, plain images).}
\label{tab:app-total}
\begin{center}
\small
\begin{tabular}{lcccccc}
\multicolumn{1}{c}{\bf SHAPES IN IMAGE} &1--10 &11--20 &21--30 &31--40 &41--50 &All
\\ \hline \\
Correct &38 / 38 &92 / 92 &44 / 45 &18 / 22 &1 / 3 &193 / 200 \\
\end{tabular}
\end{center}
\end{table}

\paragraph{Recognizing single shapes.} For each scene, we crop one shape from the plain image and ask whether it belongs to the scene's target class, with the same inclusion rules as in the counting prompt. For example, the scene question ``How many parallelograms are there?'' becomes ``Is this shape a parallelogram?''. If the scene contains the target class, we crop a random target shape (answer Yes); otherwise, we crop a random shape (answer No). The model answers correctly for 95.9\%. 

\paragraph{Summary.} The model counts all shapes correctly in 96.5\% of scenes and recognizes the single shape in 95.9\%, but counts the target class correctly in only 72.0\%. Most counting errors therefore do not come from missing or misreading individual shapes. 

\section{ReGap-Bench Details}
\label{app:bench}

All scenes are generated in Python with Pillow from a fixed random seed. For each scene, we first sample a specification (the items, their geometry, and the question) and then draw every condition from that same specification, so item positions, sizes, and orientations are identical across conditions. The gold answer is computed from the specification.

\subsection{Dataset Statistics}
\label{app:stats}

Table~\ref{tab:app-stats} summarizes the benchmark. The 1{,}000 scenes yield 3{,}800 images.

\begin{table}[h]
\caption{ReGap-Bench statistics. Each task has 200 scenes. Layout is the number of rows $\times$ columns of cells. Means are given in parentheses.}
\label{tab:app-stats}
\centering
\small
\setlength{\tabcolsep}{3.5pt}
\begin{tabular*}{\linewidth}{@{\extracolsep{\fill}}lccccr@{}}
\toprule
\textbf{Task} & \textbf{Layout} & \textbf{Items} & \textbf{Answer} & \textbf{Image size (px)} & \textbf{Images} \\
\midrule
Shape counting & 3--7 $\times$ 3--7 & 5--47 (19.0)  & 0--22 (3.2)   & 490--1130 $\times$ 580--1340 & 800 \\
Angle ranking  & 1--3 $\times$ 3--9 & 3--23 (10.4)  & Ranking       & 490--1450 $\times$ 170--490  & 600 \\
Size ranking   & 1--3 $\times$ 5--7 & 4--13 (7.7)   & Ranking       & 1280--1790 $\times$ 260--770 & 600 \\
Clock ranking  & 1--3 $\times$ 3--6 & 3--18 (9.3)   & Ranking       & 490--970 $\times$ 170--490   & 1{,}200 \\
Transformation & 1 + 4 panels       & 1 shape       & Option (A--D) & 660 $\times$ 1000            & 600 \\
\bottomrule
\end{tabular*}
\end{table}

\subsection{Task Generation}
\label{app:tasks}

We pre-define shapes from 11 classes: square, rectangle, trapezium, parallelogram, regular pentagon, equilateral triangle, isosceles triangle, right triangle, right isosceles triangle, circle, and ellipse. 

\paragraph{Shape counting.} We randomly sample shapes from the 11 classes to fill the cells. Some cells are left empty. We add marks as follows: squares and rectangles have right-angle marks at all corners; squares, pentagons, and equilateral triangles have one tick on every side; rectangles and parallelograms have one tick on one pair of opposite sides and two ticks on the other; isosceles triangles have one tick on each equal side; right triangles have a right-angle mark; right isosceles triangles have a right-angle mark and one tick on each leg; circles have two dotted radii, each with one tick. Trapezia and ellipses have no marks. Each shape has a random orientation and a random size, and classes with free proportions (e.g., rectangles, isosceles triangles, ellipses) also get random proportions.  The target class is drawn from all 11 classes, not only those present, so the answer can be zero. The gold count follows class inclusion: a square also counts as a rectangle, parallelogram, and trapezium; a rectangle also counts as a parallelogram and trapezium; a parallelogram also counts as a trapezium; equilateral and right isosceles triangles also count as isosceles triangles; a right isosceles triangle also counts as a right triangle; and a circle also counts as an ellipse. The question uses one of five templates, such as ``How many parallelograms are there?'' and ``What is the total number of circles in the image?''

\paragraph{Angle ranking.} Angles are whole degrees between $5^\circ$ and $355^\circ$, and any two angles in a scene differ by at least $5^\circ$; unused cells are left empty. All angles in a scene has a ray that share the same direction ($0^\circ$, $90^\circ$, $180^\circ$, or $270^\circ$) and the turning direction (clockwise or anticlockwise). The ranking direction is drawn at random for each scene.

\paragraph{Size ranking.} All shapes in a scene belong to one of the 11 classes above, drawn without marks, and share the same rotation ($90^\circ$, $180^\circ$, $270^\circ$, or $360^\circ$) and proportions (aspect ratio between 1.2 and 2.2 for classes with free proportions). Each shape has a scale factor between 0.10 and 0.95 that sets its linear size, and any two scale factors in a scene differ by at least 0.05. The ranking direction is drawn at random for each scene.

\paragraph{Clock ranking.} Times are drawn at random from the 1{,}440 minutes of the day, and any two clocks in a scene differ by at least 5 minutes. The hour hand moves with the minutes (e.g., at 2:30 it points halfway between 2 and 3). The ranking direction is drawn at random for each scene.

\paragraph{Transformation prediction.} Each image has five panels: the source panel (O) at the top and four options below it in a $2 \times 2$ layout. The source shape is one of the 11 classes, drawn without marks near the panel center, and $P$ is placed near the center. Each chain has three steps. A step is one of: a rotation about $P$ (90$^\circ$ clockwise or anticlockwise, 180$^\circ$, or 270$^\circ$ clockwise); a dilation about $P$ by 0.5 or 2; a reflection across a horizontal or vertical line $XY$ at a random position, used at most once per chain; or a translation up, down, left, or right, used only as the last step. We discard chains that move the shape outside the panel. The point $P$ and the line $XY$ appear in all five panels. An example question is ``(O) is rotated 90 degrees clockwise around $P$, then reflected across $XY$, then translated left. Which option shows the final result?'' Wrong options are chosen at random from a pool of faulty chains: one step changed to a similar but wrong transformation (e.g., a rotation by a different angle or about the shape's center, a reflection across a different line, a dilation with a different scale or center, or a translation in a different direction), one step removed, one step applied alone, or the steps applied in reverse order. If fewer than three valid options remain, we add a random single step. We reject any option that looks nearly the same as another (mean pixel difference below 3 on a 0--255 scale). Because the question does not state how far a translation moves the shape, we also reject any option that shows the correct shape moved in the same direction.

\subsection{Reference Conditions}
\label{app:references}

\paragraph{Numbering.} In shape counting, a label ``($k$)'' is printed above each shape, numbered row by row across all shapes. In clock ranking, the numbers 1--12 are printed inside the dial at the hour ticks; the two incomplete versions print only (3, 6, 9, 12), or (6, 12).

\paragraph{Shading.} Shading fills interiors with light gray instead of white; outlines are unchanged. It fills every shape in shape counting and size ranking, the region between the two rays indicated by the red arc in angle ranking, the whole dial face in clock ranking, and the shape in all five panels in transformation prediction.

\paragraph{Grid lines.} In shape counting and the three ranking tasks, light gray lines are drawn every 15 px. In transformation prediction, lines are drawn every 30 px in each $300 \times 300$ px panel; a translation moves the shape by exactly two grid cells.

\section{Evaluation Details}
\label{app:eval}

\lstdefinestyle{prompt}{
  basicstyle=\small\ttfamily,
  breaklines=true,
  breakindent=0pt,
  columns=fullflexible,
  frame=single,
  framesep=6pt,
  xleftmargin=6pt,
  xrightmargin=6pt,
}

\subsection{Prompts} 
\label{app:prompts}
Each task uses a single prompt template, shown below. The placeholder \texttt{\{question\}} is replaced by the question for each sample, e.g., ``How many parallelograms are there?'' for shape counting or ``Rank the clocks from earliest to latest by their labels.'' for clock ranking. The prompt depends only on the task and the question: within each paired set, all reference conditions receive identical text, and only the image differs.

\textbf{Shape counting.}
\begin{lstlisting}[style=prompt]
{question}

**Rules**:
- Rectangles include squares.
- Parallelograms include rectangles.
- Trapeziums include parallelograms.
- Isosceles triangles include equilateral triangles.
- Ellipses include circles.

**Instructions**:
You may think step-by-step and show your reasoning to carefully count the shapes.
Once you have finished your reasoning, you MUST enclose your final numerical integer count within <answer> tags.

Example format:
[Your reasoning and step-by-step counting here...]
<answer>7</answer>
\end{lstlisting}

\textbf{Angle ranking.}
\begin{lstlisting}[style=prompt]
{question}

**Instructions**:
You are tasked with ranking the angles indicated by the arcs based on their visual size.
The angles are identified by numerical labels in parentheses, such as (1), (2), (3), etc.

You may think step-by-step and show your reasoning to carefully compare the angles.
Once you have finished your reasoning, you MUST enclose your final answer within <answer> tags as a standard array of integers representing the labels. Do not include the parentheses in your final array.

Example format:
[Your reasoning and step-by-step comparison here...]
<answer>[3, 1, 2, 4]</answer>
\end{lstlisting}

\textbf{Size ranking.}
\begin{lstlisting}[style=prompt]
{question}

**Instructions**:
You are tasked with ranking the shapes based on their visual size.
The shapes are identified by numerical labels in parentheses, such as (1), (2), (3), etc.

You may think step-by-step and show your reasoning to carefully compare the sizes.
Once you have finished your reasoning, you MUST enclose your final answer within <answer> tags as a standard array of integers representing the labels. Do not include the parentheses in your final array.

Example format:
[Your reasoning and step-by-step comparison here...]
<answer>[3, 1, 2, 4]</answer>
\end{lstlisting}

\textbf{Clock ranking.}
\begin{lstlisting}[style=prompt]
{question}

**Instructions**:
You are tasked with ranking the clocks based on the times shown.
The clocks are identified by numerical labels in parentheses, such as (1), (2), (3), etc.
Use the AM/PM marker when comparing times.

You may think step-by-step and show your reasoning to carefully compare the times.
Once you have finished your reasoning, you MUST enclose your final answer within <answer> tags as a standard array of integers representing the labels. Do not include the parentheses in your final array.

Example format:
[Your reasoning and step-by-step comparison here...]
<answer>[3, 1, 2, 4]</answer>
\end{lstlisting}

\textbf{Transformation prediction.}
\begin{lstlisting}[style=prompt]
{question}

**Instructions**:
You may think step-by-step and show your reasoning to carefully apply the transformations in order.
Once you have finished your reasoning, you MUST enclose your final answer within <answer> tags as a single capital letter.

Example format:
[Your reasoning and step-by-step transformation here...]
<answer>B</answer>
\end{lstlisting}

\subsection{Reference-strategy annotation}
\label{sec:ref-strategy}
We manually assign each shape-counting response of Gemini-3.1-Flash-Lite and GPT-5.4-mini on plain and numbered images (800 responses) to one of four categories: (A) numbered reference, which uses object numbers as addresses, such as ``(3)'' or ``shape 7''; (B) row-and-column reference, which uses a systematic row, column, or cell scheme, such as ``row 2, column 3''; (C) vague spatial reference, which uses only phrases such as ``top left'' without a stable scheme; and (D) unclear or no reference, which gives the answer directly or cannot be classified. We call categories A and B \emph{systematic references}. We call categories C and D \emph{ambiguous} references.

\subsection{Pairwise accuracy} A ranking of $n$ items contains $n(n-1)/2$ item pairs. For a valid predicted ranking, pairwise accuracy is the fraction of these pairs placed in the same order as in the gold ranking; it equals $(\tau + 1)/2$, where $\tau$ is Kendall's rank correlation. We average it over the scenes with a valid ranking. An answer is invalid if it cannot be parsed or does not list every item label exactly once.

\subsection{Wrong-side readings in angle ranking} 
\label{app:wrong-side}
We use only angles below $90^\circ$ or above $270^\circ$. For such an angle with true value $\theta$, the wrong-side value $360^\circ - \theta$ lies on the other side of $180^\circ$ and differs from $\theta$ by more than $180^\circ$, so the two readings put the angle in very different places in the ranking. Within the angle's scene, let $r$ be its correct position in ascending order, that is, the number of other angles smaller than $\theta$, and let $r'$ be the position it would take if read as $360^\circ - \theta$, that is, the number of other angles smaller than $360^\circ - \theta$. Let $\hat{r}$ be its position in the predicted ranking, also in ascending order. We count a wrong-side reading if the angle is placed closer to $r'$ than to $r$, that is, $|\hat{r} - r'| < |\hat{r} - r|$. We skip angles with $r = r'$, for which the two readings give the same position, and use only scenes whose rankings are valid for both the plain and the shaded image. Because the positions are counted within each scene, the rule needs no threshold and adapts to the number of angles in the scene. As a reference, we apply the same rule to 200 random rankings of each scene.

\subsection{Grid-use judgment}
\label{app:grid-judge} 
For transformation prediction, an LLM judge (Claude Sonnet 5) reads each response of Gemini-3.1-Flash-Lite and GPT-5.4-mini with grid lines (400 responses) and decides whether it uses the grid for measurement. A response does so if it gives at least one position, distance, or size in grid cells, units, or squares, or as grid coordinates, for the shape, its vertices, the point $P$, the line $XY$, or the shape in an option panel. Rough or tentative measurements count. Coordinates count only if they describe positions in the image, not symbolic rules such as ``$(x, y)$ becomes $(-y, x)$'', and numbers that only restate the question, such as rotation angles or dilation factors, do not count. The judge sees the question and the response but not the image, and it quotes the part of the response that supports each positive decision; we checked that every such quote is supported by the response. We call a response that measures with the grid a systematic reference and any other response an ambiguous reference. The judge prompt is shown below; \texttt{\{shape\_class\}}, \texttt{\{question\}}, and \texttt{\{response\}} are filled in for each response.

\begin{lstlisting}[style=prompt, basicstyle=\footnotesize\ttfamily]
You are analyzing a model's response to a geometric transformation task. The model saw
an image with a source panel (O) showing a {shape_class} and a marked point P (and a
line XY if a reflection is involved), plus four option panels labeled A to D. All five
panels have evenly spaced light gray grid lines. The model was asked: "{question}".
You see only the model's written response, not the image.

Decide whether the response uses the grid to measure. Choose exactly one:

(A) Uses the grid for measurement: the response gives at least one position, distance,
    or size in grid cells, units, or squares, or as grid coordinates, for the shape,
    its vertices, the point P, the line XY, or the shape in an option panel. Examples:
    "the rectangle is 2 units wide and 3 units tall", "P is at (0, 0) and the top
    vertex is at (1, 2)", "the shape moves 2 cells to the left", "in option B the
    shape is 4 squares from P".
(B) Does not use the grid for measurement: the response does not mention the grid, or
    mentions it only in general terms without giving any number of cells or grid
    coordinates. Examples: "the shape moves left on the grid", "P stays in the same
    place on the grid", "option C matches the expected position".

Rules:
1. Rough or tentative measurements count as (A).
2. Coordinates count as (A) only if they describe positions in the image, such as
   positions read from the grid or measured relative to P. Symbolic rules applied to
   unspecified points, such as "(x, y) becomes (-y, x)", do not count by themselves,
   and neither does only setting P as the origin.
3. Numbers that only restate the question, such as rotation angles, dilation factors,
   or the number of steps, do not count.
4. Only grid cells, units, squares, or grid coordinates count; pixels and vague words
   such as "slightly" do not.
5. A measurement counts even if it appears in a heading or on a different line from
   the shape's name.

Return {"grid_use": "A" or "B", "evidence": "a short exact quote supporting your
choice, or an empty string if there is none"}

Response:
{response}
\end{lstlisting}

\section{Other Results}
\label{app:results}

 Table~\ref{tab:app-pairwise} lists pairwise accuracy for angle and size ranking, Table~\ref{tab:app-wrongside} gives the wrong-side analysis of angle ranking. Table~\ref{tab:app-sizegap} splits size-ranking pairwise accuracy by size gap.

\begin{table}[h]
\caption{Pairwise accuracy (\%) on angle and size ranking, averaged over valid rankings.}
\label{tab:app-pairwise}
\begin{center}
\small
\setlength{\tabcolsep}{4pt}
\begin{tabular}{lccccc}
 &\multicolumn{2}{c}{\bf ANGLE RANKING} &\multicolumn{3}{c}{\bf SIZE RANKING} \\
\multicolumn{1}{c}{\bf MODEL} &Plain &Shading &Plain &Shading &Grid lines
\\ \hline \\
Claude-Opus-4.7          &88.5 &90.9 &96.8 &97.4 &97.6 \\
Gemini-3.1-Flash-Lite    &76.4 &85.2 &95.3 &96.2 &95.8 \\
GPT-5.4-mini             &80.1 &80.5 &96.2 &96.7 &96.6 \\
Qwen3-VL-30B-A3B         &59.7 &68.0 &92.4 &93.2 &93.1 \\
InternVL3.5-38B          &69.2 &76.1 &91.1 &93.6 &91.4 \\
\end{tabular}
\end{center}
\end{table}

\begin{table}[h]
\caption{Wrong-side readings in angle ranking: the share (\%) of angles below $90^\circ$ or above $270^\circ$ that are placed closer to their wrong-side position than to their true position (Appendix~\ref{app:eval}). Scenes are those with valid rankings for both the plain and the shaded image. Random is the rate for random rankings of the same scenes.}
\label{tab:app-wrongside}
\centering
\small
\setlength{\tabcolsep}{4pt}
\begin{tabular}{lccccc}
\toprule
& & & \multicolumn{3}{c}{\textbf{Wrong-side rate (\%)}} \\
\cmidrule(lr){4-6}
\textbf{Model} & \textbf{Scenes} & \textbf{Angles} & Plain & Shading & Random \\
\midrule
Claude-Opus-4.7       & 200 & 1{,}030 & 3.3  & 1.7  & 47.8 \\
Gemini-3.1-Flash-Lite & 186 & 925     & 18.4 & 8.5  & 47.8 \\
GPT-5.4-mini          & 181 & 914     & 13.3 & 13.8 & 47.8 \\
Qwen3-VL-30B-A3B      & 95  & 377     & 31.6 & 27.9 & 47.0 \\
InternVL3.5-38B       & 161 & 716     & 23.9 & 13.1 & 47.7 \\
\bottomrule
\end{tabular}
\end{table}

\end{document}